\documentclass[acmtog, screen, nonacm]{acmart} 
\usepackage{tcolorbox}
\tcbuselibrary{skins,breakable}
\usepackage{enumitem}
\usepackage{xcolor}
\usepackage{todonotes}

\definecolor{Accent}{HTML}{2D7FF9}  
\definecolor{Panel}{HTML}{F8F9FB}   
\definecolor{Frame}{HTML}{E4E7EB}   
\definecolor{Ink}{HTML}{111111}     

\newtcolorbox{policybox}[2][]{%
  enhanced,
  breakable,
  colback=Panel,
  colframe=Frame,
  coltitle=Ink,
  title={\sffamily\bfseries #2},
  fonttitle=\sffamily\bfseries,
  arc=2mm,
  boxrule=0.4pt,
  left=2mm,right=2mm,top=1.5mm,bottom=1.5mm,
  boxsep=1mm,
  borderline west={2pt}{0pt}{Accent},
  before skip=6pt, after skip=6pt,
  sharp corners,
  drop shadow={black!10!white},
  #1
}

\AtBeginDocument{%
  }

\begin{document}

\title{Rethinking Science in the Age of Artificial Intelligence}

\author{Maksim E. Eren}
\email{maksim@lanl.gov}
\orcid{0000-0002-4362-0256}
\affiliation{%
  \institution{Information Systems and Modeling, Los Alamos National Laboratory}
  \city{Los Alamos}
  \state{NM}
  \country{USA}
}

\author{Dorianis M. Perez}
\email{dorianisp@lanl.gov}
\orcid{0000-0003-3538-8386}
\affiliation{%
  \institution{Verification and Analysis, Los Alamos National Laboratory}
  \city{Los Alamos}
  \state{NM}
  \country{USA}
}


\begin{abstract}
Artificial intelligence (AI) is reshaping how research is conceived, conducted, and communicated across fields from chemistry to biomedicine. This commentary examines how AI is transforming the research workflow. AI systems now help researchers manage the information deluge, filtering the literature, surfacing cross-disciplinary links for ideas and collaborations, generating hypotheses, and designing and executing experiments. These developments mark a shift from AI as a mere computational tool to AI as an active collaborator in science.
Yet this transformation demands thoughtful integration and governance. We argue that at this time AI must augment but not replace human judgment in academic workflows such as peer review, ethical evaluation, and validation of results. This paper calls for the deliberate adoption of AI within the scientific practice through policies that promote transparency, reproducibility, and accountability.

\end{abstract}




\maketitle



\section{Introduction}
\label{sec:introduction}

Artificial Intelligence (AI) is already delivering measurable benefits across scientific domains from materials discovery and chemistry \cite{darvish2025organa,m2024augmenting} to genomics \cite{dalla2025nucleotide} and biomedicine \cite{gao2024empowering,kim2024mdagents,tang2023medagents}. These domain specific gains unfold amid an escalating information deluge where the scientific literature is growing so quickly that maintaining a comprehensive and connected view of any field is increasingly difficult \cite{bornmann2021growth,gu2024interesting}. In response, AI is reshaping the research lifecycle itself by helping researchers triage and synthesize literature, uncover cross-disciplinary relationships, generate and test hypotheses, and design experiments, thereby accelerating the conduct of science \cite{zheng2024disciplink,yang2023large,sourati2023accelerating}.

Prior studies have surveyed distinct facets of AI for science. Wang et~al.\ synthesize how modern AI methods, including self-supervised, generative, and reinforcement learning models, now touch the full research arc, emphasizing cross-domain case studies and challenges in data stewardship, reliability, and evaluation \cite{wang2023scientific}. Krenn et~al.\ emphasize the distinction between AI prediction and understanding and position AI as a computational microscope, a resource of inspiration, or a potential agent of understanding, with corresponding epistemic criteria \cite{krenn2022scientific}. Gridach et~al.\ survey agentic systems built for scientific discovery, organizing single and multi agent architectures by level of autonomy and interaction, and cataloging tools for ideation, literature review, planning, and experiment execution while considering issues in calibration, reproducibility, and governance \cite{gridach2025agentic}. Here with this commentary, we offer a complementary, workflow-centric analysis that follows how AI reshapes the research lifecycle from ideation and collaborator discovery to forecasting, experimentation, and the psychology of discovery, while drawing on mechanisms from adjacent domains, to inform policy design.

This \emph{commentary} argues that AI is not merely accelerating scientific progress but fundamentally transforming how science itself is conducted.  Across the research lifecycle from hypothesis generation to experimental design, execution, and evaluation AI systems are becoming integral collaborators in discovery. Knowledge Graph (KG) and retrieval-augmented large language models (LLMs) driven models can reveal connections that were previously invisible \cite{gu2024forecasting,gu2024interesting}, while autonomous and human-in-the-loop systems are used for experimentation \cite{darvish2025organa, prince2024opportunities}. On the other hand, recent studies highlight the limitations of reasoning models and call for explainability, underscoring the need for governance, transparency, and evaluation frameworks to ensure trustworthiness \cite{shojaee2025illusion, gridach2025agentic, Salloum2024, lietal2023, Kauretal2022}. Together, these shifts move AI from assistant to collaborator, positioned to augment creativity and reshape scientific norms, ethics, and workflows. Here we translate our observations into policy recommendations to align scientific incentives with trustworthy AI integration. To our knowledge, these recommendations have not been addressed in prior global policy memos or position papers on AI in science \cite{gursoyAIstrategy, VEALE_2020, brief2021us, Helregel2025, Vydra2019, daMota2024}.

\section{AI and Science}
\label{sec:ai_science}

We provide a review of AI for science with a focus on the scientific workflow spanning (i) literature navigation, (ii) team formation, (iii) forecasting and early signals, (iv) hypothesis generation, (v) agentic experimentation, (vi) realistic evaluation, and (vii) psychology and AI parallels. We draw cross-domain lessons and provide opinionated guidance on system design and governance.

\subsection{Navigating the Deluge of Papers}
Modern science is saturated with rapidly growing and fragmented literature where the bottleneck is no longer access but sense making. AI can serve as a literature navigation partner for research papers. In Systematic Literature Reviews (SLRs), AI can help frame problem spaces, diversify and reformulate search queries, cluster and triage long result, translate jargon across fields, and maintain provenance with explicit citations \cite{bolanos2024artificial, wagnerlitreview2022}. A variety of tools using this framework and others have been posed in the literature spanning all scientific disciplines \cite{bolanos2024artificial, oyelude2024, Danler2024Quality}. Tools like \textit{DiscipLink} perform interdisciplinary searches by proposing exploratory questions, expanding discipline specific terminology, structuring results into thematic clusters, and annotating "information scent," while keeping the human in control to steer, prune, and collect \cite{zheng2024disciplink}. Further, LLMs improve scholarly search by generating research intent queries and enabling retrieval-augmented (RAG) pipelines with transparent grounding and systematic query expansion \cite{ajith2024litsearch}.

\subsection{Identifying Collaborators and Forming Teams}
Beyond literature search, LLMs can help identify collaborators, which is a critical function in science, where breakthroughs often hinge on mixing expertise across domains \cite{Vallurucollab}. Human-aware forecasting maps author and concept pathways on hypergraphs and uses an expert density control to distinguish well staffed intersections from "alien" gaps highlighting where cross-field teaming could unlock novelty \cite{sourati2023accelerating}. KG plus LLM ideation personalizes suggestions by analyzing researchers' publication histories and proposing collaborator and concept pairings, and project sketches aligned to complementary expertise \cite{gu2024interesting}. Agentic, citation graph driven workflows iterate over the literature while preserving traceable context, supporting judgments about fit and division of labor \cite{baek2404researchagent}. Mixed-initiative tools (e.g., \textit{DiscipLink}) expand and structure scientific searches, surfacing adjacent communities and potential collaborators \cite{zheng2024disciplink}. Complementing this, instruction following multimodal web agents (e.g., \emph{WebGUM}) can compile open-web evidence lab pages, code, datasets, and talks by jointly reasoning over screenshots and HTML, that could provide inputs to such systems \cite{furuta2023multimodal}. Analogously, intent conditioned, strategically grounded dialogue can be repurposed to broker collaborations to model collaborator' beliefs and goals to negotiate roles, align constraints, and draft trust-aware, actionable teaming plans beyond simple match making \cite{meta2022human}. Likewise, theory-of-mind aware planning could improve teaming by inferring collaborators' latent goals, constraints, and beliefs from partial signals and past interactions, then adapting outreach, roles, and timelines to preempt coordination failures under uncertainty \cite{guo2023suspicion}.

\subsection{Forecasting Science and Hypothesis Generation}
Beyond search, AI now helps anticipate where science is heading and supports the turning these signals to concrete ideas. On the forecasting side, evolving KGs built from concept co-occurrences in titles and abstracts, augmented with citation derived impact signals and temporal trends, enable link-\ prediction models that flag unseen concept pairs that are likely to become high impact \cite{gu2024forecasting}. Human-aware formulations enrich these graphs with author structure, using author mediated path probabilities and an expert density control to weight who can plausibly connect ideas and to trade off near term feasibility against longer horizon "alien" directions \cite{sourati2023accelerating}.

Complementary systems operationalize hypothesis generation with literature grounding and iterative critique LLM agents. \emph{SCIMON} combines retrieval based inspiration with contrastive, revision oriented prompting to boost novelty while staying anchored in prior work \cite{wang2023scimon}. \emph{ResearchAgent} mimics scholarly workflow by traversing citation graphs, maintaining entity centric context stores, and using reviewer style agents to refine problem statements, methods, and experimental plans \cite{baek2404researchagent}. Open domain pipelines such as \emph{TOMATO/MOOSE} decompose hypothesis discovery into retriever–proposer–checker loops with self-feedback to improve validity, novelty, and clarity \cite{yang2023large}. Finally, KG anchored LLM approaches can rank and refine ideas judged "interesting" by domain experts, including personalized suggestions tied to researchers' publication histories \cite{gu2024interesting}. Zero-shot prompting can surface plausible directions, yet retrieval grounding and reflective self-improvement can raise novelty, validity, and specificity \cite{qi2023large,peng2023check,madaan2023self}. This extends early idea drafting systems into agentic, self-critiquing workflows that read, propose, and revise, thereby bridging forecasting signals to testable hypotheses \cite{wang2019paperrobot,li2024chain,xu2023exploring}. While recent methods use prompting to organize review and critique style agents, \emph{DSPy} argues for replacing brittle templates with declarative, learnable programs and optimizer tuned modules pushing the field toward reproducible, composable, and data driven alternatives to manual prompting \cite{khattab2024dspy}.

\subsection{Agentic Workflows for Experimentation}
Recent work also casts AI as an active workflow participant that "senses", "plans", "acts", and "learns" across the research cycle \cite{gridach2025agentic}. In these agentic frameworks, a planner decomposes goals into tasks; tool using executors call retrieval, simulation, or lab control APIs; critics and verifiers check assumptions and outputs; and shared memory tracks state, decisions, and open questions. The result is not a single model call but an orchestrated sense–plan–act–reflect loop that persists through literature review, hypothesis generation, study design, execution, and analysis.

Role structured, multi-agent designs make this loop reliable at scale. Frameworks like \emph{MetaGPT} encode Standard Operating Procedures (SOPs) so that specialized agents (e.g., planner, researcher, critic, engineer) produce standardized artifacts (i.e. requirements, study protocols, code, and reports) that enable precise handoffs and reduce conversational drift \cite{hong2023metagpt}. Complementary approaches such as \emph{CAMEL} formalize role-play and negotiation between agents to improve decomposition, constraint handling, and iterative refinement \cite{li2023camel}. Agent orchestration also extends from reasoning to acting in physical laboratories. Agentic platforms connect literature, planning, code execution, and robotics to design and run experiments with minimal supervision where co-scientist demonstrates autonomous planning across documentation navigation, instrument control, and reaction optimization \cite{boiko2023autonomous}; \emph{ORGANA} integrates natural language and human-in-the-loop control for diverse wet-lab procedures \cite{darvish2025organa}; and \emph{LLaMP} uses hierarchical agents, data sources, and simulation tools to support property prediction, structure manipulation, and synthesis reasoning with response consistency tracking \cite{chiang2024llamp}. Chemistry toolchains like \emph{ChemCrow} further extend LLMs with domain specific APIs and evaluators \cite{m2024augmenting}. 



\subsection{Realistic Benchmarks and Evaluation}
Evaluating agentic systems under the constraints they face in practice remains an open problem. End-to-end ML engineering testbeds expose contamination, brittle scaling with compute, and weak generalization beyond memorized patterns, which highlights the need for reproducible setups with strict data hygiene \cite{chan2024mle,huang2023mlagentbench,kang2024researcharena}. In literature retrieval, author written intents and deliberately hard queries stress reasoning over full texts rather than lexical shortcuts, pushing evaluation toward provenance aware RAG behavior \cite{ajith2024litsearch}. Community studies of LLM judges show inconsistency and bias in preference based scoring, so automated assessors should be complemented by human grounded protocols and blinded, rubric driven reviews \cite{zheng2023judging,chiang2024chatbot}.

Beyond task accuracy, realistic evaluation should capture process metrics, including adherence to provenance, claim level citation correctness, tool call success rates, handoff quality between agents, calibration of confidence, and time or cost to solution under API failures, shifting corpora, and strict latency or compute budgets \cite{gridach2025agentic}. Claim grounding can be tested directly with citation faithfulness checks such as \emph{CiteME} \cite{press2024citeme}, while response reliability can be summarized with consistency metrics like \emph{SCoR} \cite{chiang2024llamp}. Complexity controlled stress tests help reveal collapse regimes in reasoning and should be part of any agent evaluation suite \cite{shojaee2025illusion}. Since scientists work in teams, benchmarks should also report human plus AI complementarity and appropriate trust calibration, not only solo agent scores \cite{bansal2021does,ma2023should}. Finally, facility and lab tracks should simulate safety checks, tool faults, and closed-loop control to reflect real scientific environments \cite{laurent2024lab,prince2024opportunities}.

\subsection{Psychology and AI Parallels in Scientific Discovery}
Chadwick and Segall frame scientific judgment as creative yet fallible, highlighting overconfidence, premature closure, availability and recency biases, and poor calibration, while noting that calibrated optimism can sustain effort and widen search under uncertainty. They recommend countermeasures that pair exploration with evidence based workflows, multi-attribute reasoning, and practice settings with rapid feedback to recalibrate judgment without dampening creativity \cite{chadwick2010overcoming}. May be seen as design cues for AI for science, these lessons can map to recent systems. To resist premature closure, hypothesis pipelines that decompose, critique, and revise act as commitment brakes. \emph{SCIMON} retrieves inspirations, then uses contrastive, revision oriented prompting to boost novelty while staying grounded in prior work \cite{wang2023scimon}. \emph{ResearchAgent} traverses citation graphs, maintains entity centric context, and employs reviewer style agents for multi-attribute deliberation and adversarial critique \cite{baek2404researchagent}. Open domain \emph{TOMATO/MOOSE} splits discovery into retriever–proposer–checker loops with self-feedback, providing fast practice cycles before commitment \cite{yang2023large}. To counter availability and recency effects, mixed-initiative literature tools and hard retrieval benchmarks emphasize explicit provenance, targeted query expansion, and cluster structure while keeping humans in the loop \cite{zheng2024disciplink,ajith2024litsearch}. To balance optimism with realism, human-aware re-ranking with author pathway density offers a tunable trade-off between feasibility and novelty, surfacing near term, high capacity links or longer horizon "alien" directions \cite{sourati2023accelerating}. At the systems level, role structured multi-agent frameworks encode standard operating procedures and audit trails for proposing, critiquing, retrieval, and execution, echoing evidence based workflows and checklists \cite{hong2023metagpt,gridach2025agentic}.

\section{Policy Recommendations}
\label{sec:policy_recommendations}


Building on these findings, this section translates the literature's strengths and failure modes into policy recommendations.

\subsection{Fund Responsible and Interdisciplinary Integration of AI into Scientific Workflows}
AI now influences every stage of the research cycle from literature discovery to experimental execution, yet many systems remain opaque or narrowly optimized. Public investment should therefore prioritize \emph{open, interpretable, and auditable} tooling and \emph{mixed-initiative} designs that preserve human agency while enabling responsible automation. Literature co-exploration systems (e.g., \emph{DiscipLink} \cite{zheng2024disciplink} and \emph{LitSearch} \cite{ajith2024litsearch}) illustrate how human judgment coupled with provenance-aware retrieval can outperform full automation, while laboratory frameworks such as \emph{Coscientist} and \emph{ORGANA} demonstrate the value of human-in-the-loop experimentation \cite{boiko2023autonomous,darvish2025organa}. Domain toolchains that integrate \emph{consistency and calibration} signals (e.g., SCoR) make reasoning more inspectable \cite{chiang2024llamp}.

Complementing these technical priorities, funding agencies should also support \emph{interdisciplinary fellowships} connecting computer science, philosophy, technology, and domain science to examine how AI reshapes explanation, evidence, and creativity \cite{krenn2022scientific,wang2023scientific}. Such programs can co-develop epistemic norms for provenance, reproducible agent traces, and standards for evaluating AI generated reasoning.

Together, these initiatives would establish a virtuous cycle between \emph{responsible infrastructure} and \emph{reflective inquiry}, ensuring that investment in scientific AI advances both technical rigor and philosophical understanding. Funding should require shared benchmarks and standardized reporting of model versions, prompts, datasets, uncertainty, and agent logs; support open APIs and datasets; and mandate audits for contamination and generalization (cf.\ agent engineering benchmarks \cite{chan2024mle}). We argue that at this time complementarity between humans and AI, but not pure automation yields more transparent, creative, and trustworthy science.

\subsection{Establish Third-Party Oversight for Autonomous Experimentation}
As agentic systems plan and execute experiments, independent review is essential for safety, reliability, and ethics. Self-driving lab exemplars in chemistry and biomedicine highlight both promise and risk \cite{boiko2023autonomous,darvish2025organa,gao2024empowering,gridach2025agentic}. Here the recommendation is domain appropriate oversight bodies (akin to IRBs/biosafety committees) to evaluate: (i) fail-safes and intervention protocols, (ii) data provenance and access controls, (iii) dual-use risk, and (iv) calibration/consistency of model reasoning. Given the current arguments on brittleness of current reasoning models at rising complexity \cite{shojaee2025illusion}, approvals should condition deployment on human override, red-teaming, and staged roll-outs with audit logs.

\subsection{Require Transparent AI Involvement and Disclosure Standards}
To make AI's influence in science visible, reproducible, and accountable, research workflows and publication venues should require explicit disclosure of AI involvement. This includes \emph{preregistered AI involvement statements} specifying where and how AI contributed, detailing models and versions used, prompts, datasets, retrieval sources, and decision points. Mixed-initiative systems shape problem formulation and discovery trajectories \cite{zheng2024disciplink,ajith2024litsearch}; tools like \textit{CiteME} highlight provenance and citation risks when such transparency is absent \cite{press2024citeme}. Statements should link to \emph{agent traces} (e.g., proposer–critic iterations, sampling settings, temperature) and uncertainty tags, complementing data and code availability.

In parallel, clear \emph{authorship and attribution standards} are needed to delineate human and AI contributions. Consistent with emerging norms, AI systems should not be listed as authors but instead, manuscripts should include a dedicated "AI Contributions" section enumerating specific tasks (e.g., hypothesis ideation, retrieval, drafting, analysis) and identifying the responsible human supervisors. This preserves accountability while acknowledging substantive AI assistance, reflecting evolving conceptions of scientific credit in the age of AI \cite{wang2023scientific,krenn2022scientific}.

These transparency norms should extend to \emph{peer review and editorial processes}. Journals should require reviewers and editors to disclose any AI assistance used specifying system, version, and prompts, and \emph{forbid} delegating final decisions to AI. Where AI tools inform evaluation, their evidence trails (citations, retrievals, reasoning summaries) should accompany the review. Given persistent reasoning and calibration failures in current models \cite{shojaee2025illusion}, human adjudication and appeals must remain central to maintaining rigor and trust in scientific publishing.

\subsection{Integrate and Incentivize Human-AI Literacy and Collaborative Research Practices}
AI literacy should become a foundational competency for researchers, coupling technical fluency with ethical and epistemic awareness. Curricula should pair practical skills, RAG, prompting, and agent orchestration with modules on interpretability, bias, provenance, and uncertainty communication \cite{wang2023scientific,hope2023computational}. Students and researchers should learn to read agent logs, assess calibration and consistency, and practice \emph{controlled exploration} (e.g., adjusting temperature or sampling diversity) responsibly during ideation.

Beyond education, research policy should incentivize human–AI collaboration that foregrounds complementarity, oversight, and shared agency. Human-aware forecasting that exposes author pathways \cite{sourati2023accelerating}, mixed-initiative literature systems \cite{zheng2024disciplink}, and reflective biomedical agents \cite{gao2024empowering} exemplify such partnerships. Calls and reviews should explicitly value: (i) human-in-the-loop checkpoints, (ii) reflective or critic loops before commit, (iii) transparent reporting of exploration settings and downstream re-ranking, and (iv) open, auditable pipelines. Educational and incentive reforms, taken together, would cultivate literate, reflective researchers and reward designs that integrate controlled exploration, provenance tracking, and human governance, producing more creative, trustworthy science.

\section{Conclusion}
\label{sec:conclusion}

Artificial Intelligence (AI) is no longer a peripheral aid to research. Instead, AI is becoming an active collaborator that helps scientists navigate the literature deluge, generate and assess hypotheses, plan and execute experiments, and synthesize results across disciplines. The papers reviewed here show clear and measurable benefits alongside equally clear limits: current systems can be brittle, biased, and opaque as task complexity rises. The path forward, for now, is neither uncritical automation nor status-quo skepticism, but a mixed-initiative paradigm in which AI \textit{augments}, and \textit{never replaces} human judgment.

Despite significant progress, important gaps remain. The scientific community lacks standardized, auditable agent logs that enable end-to-end reproducibility across literature, ideation, and lab stages; evaluations must become more contamination resistant and reflective of real constraints; autonomous experimentation requires domain specific safety scaffolds and independent oversight; and socio-technical defaults should make complementarity human oversight, provenance, and uncertainty reporting the norm rather than an afterthought. Deliberate, targeted policies can move these from aspiration to baseline practice.


\section*{Acknowledgment}
This manuscript has been assigned LA-UR-25-30965. The funding for this paper was provided by Los Alamos National Laboratory (LANL) as part of the National Security and International Studies (NSIS) Fellowship. LANL is operated by Triad National Security, LLC, for the National Nuclear Security Administration of the U.S. Department of Energy (Contract No. 89233218CNA000001).

\bibliographystyle{ACM-Reference-Format}
\bibliography{references}

\end{document}